\documentclass[conference]{IEEEtran}
\IEEEoverridecommandlockouts
\usepackage{cite}
\usepackage{amsmath,amssymb,amsfonts}
\usepackage{algorithmic}
\usepackage{graphicx}
\usepackage{textcomp}
\usepackage{xcolor}
\usepackage{booktabs}
\usepackage{hyperref}
\usepackage{adjustbox}

\usepackage[caption=false, font=footnotesize, labelfont=sf, textfont=sf]{subfig}

\begin{document}

\title{Reliable Classification with Conformal Learning and Interval-Type 2 Fuzzy Sets}
\author{
\IEEEauthorblockN{1\textsuperscript{st} Javier Fumanal-Idocin, 2\textsuperscript{nd} Javier Andreu-Perez}
\IEEEauthorblockA{\textit{School of Computer Science and Electronic Engineering} 
\textit{University of Essex}\\
Colchester, United Kingdom \\
\{j.fumanal-idocin, j.andreu-perez\}@essex.ac.uk}
\\
}

\maketitle

\begin{abstract}
Classical machine learning classifiers tend to be overconfident can be unreliable outside of the laboratory benchmarks. Properly assessing the reliability of the output of the model per sample is instrumental for real-life scenarios where these systems are deployed. Because of this, different techniques have been employed to properly quantify the quality of prediction for a given model. These are most commonly Bayesian statistics and, more recently, conformal learning. Given a calibration set, conformal learning can produce outputs that are guaranteed to cover the target class with a desired significance level, and are more reliable than the standard confidence intervals used by Bayesian methods. In this work, we propose to use conformal learning with fuzzy rule-based systems in classification and show some metrics of their performance. Then, we discuss how the use of type 2 fuzzy sets can improve the quality of the output of the system compared to both fuzzy and crisp rules. Finally, we also discuss how the fine-tuning of the system can be adapted to improve the quality of the conformal prediction. 
\end{abstract}


\section{Introduction}
Uncertainty quantification is a key feature of modern machine learning systems, especially those deployed in real-life settings \cite{klas2018uncertainty}. Many applications require realistic performance estimations, which are not directly obtained from classical logits and predictions, both in classification and regression \cite{hullermeier2021aleatoric}. Fields such as medicine, law, and finance, where human stakeholders must make final decisions, particularly benefit from robust uncertainty quantification \cite{chua2023tackling}. Bayesian statistics have been widely utilized to quantify uncertainty in decisions, but these methods can sometimes result in overconfident intervals \cite{abdar2021review, moore2008trouble}. Consequently, Bayesian approaches have, in some cases, been replaced by conformal predictions.

Conformal learning involves using a calibration set and a non-conformal score, enabling the classifier's output to be a set instead of single prediction that contains the true value with a guaranteed coverage \cite{MAL101}. This methodology has seen extensive adoption in industrial applications, such as predictive maintenance, fraud detection, and personalized medicine, where reliable uncertainty bounds are critical for operational decisions \cite{singh2024uncertainty, luo2024sample, balasubramanian2014conformal}. Conformal prediction techniques offer a practical and mathematically rigorous framework for handling uncertainty, making them highly attractive for integration into real-world systems.

Rule-based systems are among the most popular methods in explainable AI, as they directly map input regions to specific outputs \cite{van2021evaluating}. These systems have also been extended to incorporate uncertainty quantification methods, such as fuzzy sets, to improve interpretability and reliability \cite{panda2021bayesian, fumanal2023stability}. For example, fuzzy logic systems have been employed to capture uncertainty inherent in linguistic variables or imprecise input data \cite{mendel2023explainable}. Existing fuzzy systems have been combined with Bayesian reasoning to support fuzzy states within Bayesian inference frameworks \cite{chen2022risk}. However, their integration with other uncertainty quantification methods has been limited.

Similarly, standard decision rules face challenges in accommodating conformal predictions. For example, decision trees, a widely used rule-based model, typically output binary or categorical vectors, which are not directly compatible with conformal prediction requirements. A common workaround is to use ensembles of trees, such as random forests or gradient-boosted decision trees, to provide outputs in the real $[0,1]$ range. However, this approach compromises some of the inherent interpretability advantages of individual tree models. In such cases, techniques like Shapley values are often employed to explain the outputs of ensemble models, albeit at the expense of simplicity in the explanations \cite{nguyen2021evaluation, lundberg2020local}.

In this work, we propose to use fuzzy rules with conformal learning to guarantee the coverage of the true classes using a fuzzy rule system. We show how such system works and test its performance. Then, we also show how this system can be improved by using interval type 2 fuzzy sets and different fitness functions. 

The rest of the paper goes as follows: Section \ref{sec:background} discusses the literature and previous concepts regarding conformal learning, fuzzy-rule based classification and uncertainty quantification using fuzzy sets.  Section \ref{sec:main} shows how to apply conformal learning and fuzzy rule-based classification. Subsequently, Section \ref{sec:exp} showcases the results obtained with our novel methods and compares them to other standard classification procedures. Finally, Section \ref{sec:conclusions} displays our conclusions for future work.

\section{Background} \label{sec:background}

\subsection{Conformal learning}
Conformal prediction is a framework for providing rigorous uncertainty quantification in machine learning predictions. Unlike traditional methods that often rely on distributional assumptions, conformal prediction offers distribution-free uncertainty estimates with finite-sample guarantees \cite{balasubramanian2014conformal}. At its core, the method constructs prediction sets that contain the true label with a user-specified plausibility $1-\alpha$ over repeated trials, where $\alpha$ is the desired significance level.

The fundamental principle of conformal prediction relies on the concept of nonconformity scores, which measure how different a new example is from previously observed data. Given a training set ${(X_1,Y_1),\dots,(X_n,Y_n)}$ and a nonconformity measure $s$, which is commonly: $s_i=1-\hat{f}(X_i)_{Y_i}$, we define $\hat{q}$ as the $ceil((n+1)(1-\alpha))/n$ quantile of $\{s_1,\dots,s_n\}$. For a new test point $X_{n+1}$, the prediction set $\hat{f}(X_{n+1})$ is constructed as:
\begin{equation} \label{eq:conformal_pred}
    C(X_{n+1}) = \{y : \hat{f}(X_{n+1})\ge1-\hat{q}\}.     
\end{equation}

This approach guarantees $P(Y_{n+1} \in C(X_{n+1})) \ge 1-\alpha$ under the assumption of exchangeability (when the order of the samples does not matter, i.e. no concept drift is present).

\subsection{Fuzzy rule-based classification}

Fuzzy rule-based classification involves categorizing observations into different classes using rules structured as follows \cite{kosko1986fuzzy}:

\begin{equation}
\text{IF } \mathbf{x}_1 \text{ is } \mathbf{a}_{1}, \dots, \mathbf{x}_n \text{ is } \mathbf{a}_{n}, \text{ THEN class } j,
\end{equation}.
\noindent for $j\in \{1, \dots, C\}$.
Typically, a fuzzy rule-based classifier (FRBC) evaluates the degree of match between an input and all available rules, selecting the rule that maximizes this value. Rather than using solely the rule with the highest value, multiple rules can be considered for a decision \cite{wieczynski2022applying}. However, this may reduce the interpretability of the system, as the final decision would then depend on more than one pattern simultaneously.

\subsection{Uncertainty quantification using fuzzy sets}
Fuzzy sets have been used as a means of measuring both aleatoric and epistemological uncertainty \cite{figueroa2024cosine, he2015mixed, mendel2023explainable}. The use of fuzzy probabilities has also been used to measure aleatoric uncertainty \cite{buckley2006fuzzy}. 

Fuzzy sets are a particular kind of credal set that is also used to model uncertainty \cite{hullermeier2022quantification}, especially for epistemic uncertainty \cite{sale2023volume}. Normally, these sets are used to indicate uncertainty in the predictions and to generate non-singleton outputs in both regression and classification tasks \cite{wangcredal, caprio2024credal}.

\section{Fuzzy rule-based conformal classification} \label{sec:main}

\subsection{Interval-Type 2 conformal learning}
When using interval-valued Type 2 (IV-T2) systems, we find that the predictions for each class are not numerical but an interval value. In this case, we need to use interval arithmetic to construct the conformal histogram and make a small tweak to the nonconformal score to adapt it to an interval-valued setting.

Considering that the real interval-valued prediction for the correct class is $(1,1)$, we can compute the interval-valued conformal score as $S_i=(1,1)-\hat{f}_{IV}(X_{i})_{Y_i}$. To construct the nonconformal score quantiles, we need to order the intervals. To do so, we are going to use a suitable order for intervals, which we will define using the $K_a$ operator.

Considering any closed subinterval of the unit interval $[0,1]$:
\begin{equation}
	L([0,1])=\{X=[\underline{X},\overline{X}]\mid 0\leq \underline{X}\leq \overline{X}\leq 1\}.
\end{equation}

The $K_a$ operator is defined for all $X\in L([0,1])$ and $a \in [0,1]$, by:
\begin{equation}
	K_{a}(X) = (1-a)\underline{X}+a \overline{X}.
\end{equation}   
Given the K$_a$ parameter, we can define a total order for intervals given two parameters $\alpha, \beta \in [0,1]$ with $\alpha\neq\beta$. Then, the total order $\leq_{\alpha,\beta}$, is defined using by $K_\alpha$ and $K_\beta$, for all $X,Y\in L([0,1])$, as: \cite{bustince2013generation, fumanal2021interval}

\begin{equation}
	X\leq_{\alpha,\beta} Y \hspace{0.2cm} \text{ if } \hspace{0.2cm} \begin{cases}
	K_{\alpha}(X) < K_{\alpha}(Y) \text { or }
	\\
	K_{\alpha}(X)=K_{\alpha}(Y) \text{ and }
	K_{\beta}(X) \leq K_{\beta}(Y).
	\end{cases}
\end{equation}  A total order on $L([0,1])$ is called an admissible order  \cite{bustince2013generation,fumanal2021interval} if it generalizes the standard product order between intervals, which is a partial order.

Now that we have defined the $\leq_{\alpha,\beta}$ order, we can compute the $\hat{Q}$ quantile of the {$S_1, \dots, S_n$} non-conformal scores. Then, we can reformulate Eq. (\ref{eq:conformal_pred}) as:

\begin{equation} \label{eq:conformal_pred_iv}
    C(X_{n+1}) = \{y : \hat{F}(X_{n+1})\geq_{\alpha,\beta} 1-\hat{Q}\},     
\end{equation}

\noindent which will now be suitable to use with interval-valued data. 

\subsection{Rule-wise conformal learning}

\begin{figure*}[ht]
    \centering
    \includegraphics[width=.8\linewidth]{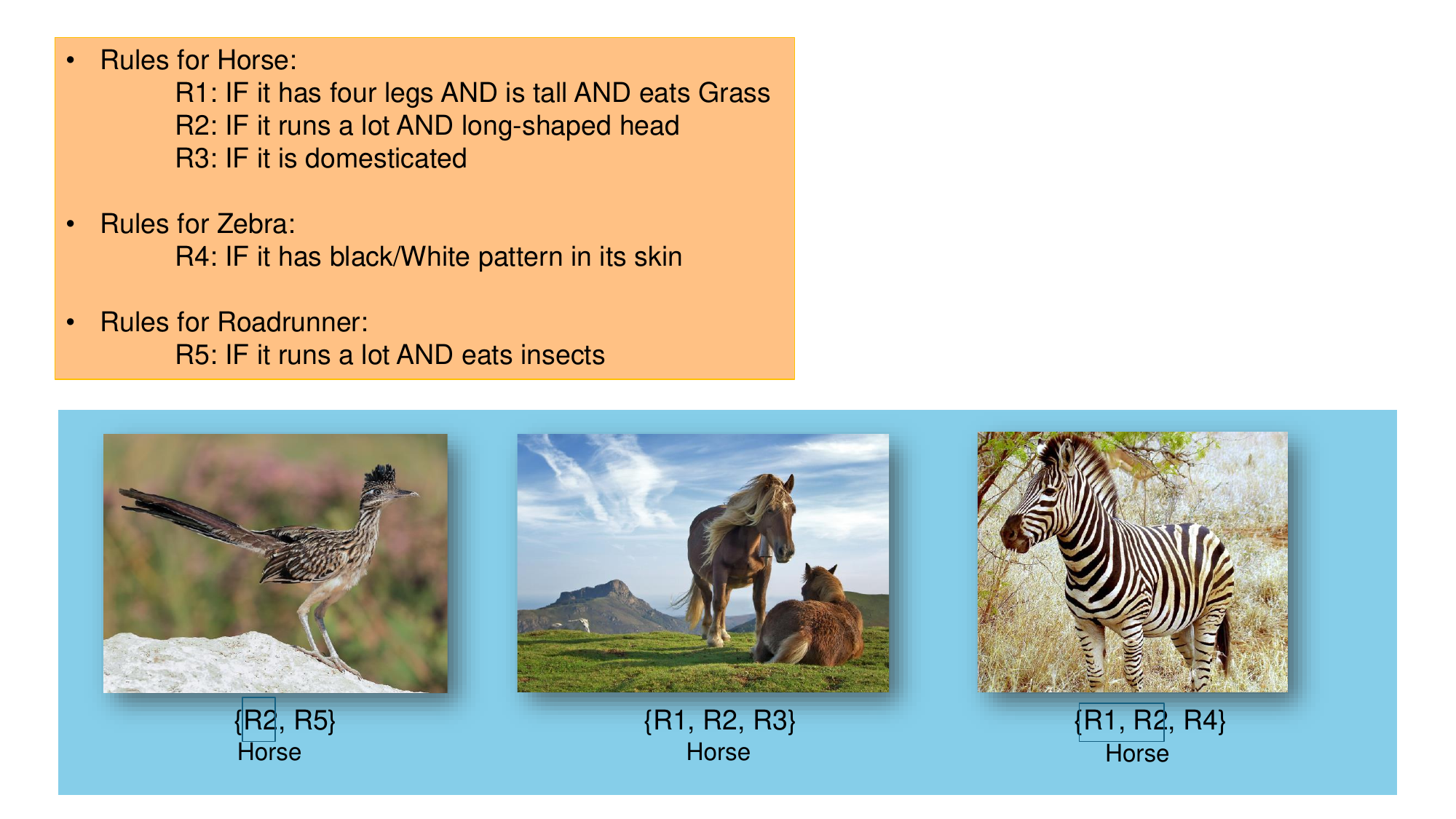}
    \caption{\textbf{Why use rule-wise predictions?.} In this example, we are working with three classes: roadrunner, horse and zebra. If we only used class-wise predictions, we would see that the horse class is always present in the predictions, but we would not know why. If we have rule-wise predictions we can identify and even amend the spurious patterns (R1 and R2 in the example) and leave those that are working fine (R3).}
    \label{fig:rulewiseconformal}
\end{figure*}

 Sometimes patterns are not reliable or complete. i.e. Figure \ref{fig:rulewiseconformal}. We can try to discard such rules using the standard quality metrics, but sometimes, they can be as deceiving as the output logits.
 
To overcome that, we can also exploit conformal learning and the fact that we can treat each rule as a single classifier, building the output in this way:
\begin{equation}
    C(X_{n+1}) = {r : \hat{f}_r(X_{n+1)} \geq 1 - \hat{q}}.    
\end{equation}
\noindent We can follow this approach by building a non-conformity histogram for each rule, but that can be prohibitively expensive if the number of rules and samples is high. However, we can build the non-conformal histograms at a classifier level and use those for each of the individual rules as well, which is considerably cheaper in terms of computing.

In this way, we can then analyse which rules are consistently giving significantly high truth degree values in incorrect cases. 

\section{Methods} \label{sec:methods}

\subsection{Evaluation and Datasets used}
In order to compute the performance, we used a list of datasets obtained from the Keel website \cite{triguero2017keel}, listed in Table \ref{tab:datasets}. Each dataset is normalised by subtracting the mean of each feature and dividing by the standard deviation.

In order to evaluate each performance, we compute the accuracy using 80/20 train-test partitions. Since these predictions are sets, we also measure the size of average the conformal set. 

\begin{table}[]
		\centering
		\caption{Datasets used for our experimental results}
		\begin{adjustbox}{max width=\linewidth}
			\begin{tabular}{cccc}
				\toprule
				Dataset & Instances & Features & Classes \\
                    \midrule
                    iris & 150 & 4 & 3 \\
                    glass & 214 & 9 & 6\\
                    haberman & 306 & 3 &  2\\
                    ionosphere & 351 & 33 & 2 \\
                    wine & 178 & 13 & 3 \\
                    balance & 625 & 4 & 3 \\
                    heart & 270 & 13 & 2 \\
                    pima & 768 & 8 & 2 \\
				\bottomrule
			\end{tabular}
		\end{adjustbox}
		\label{tab:datasets}
	\end{table}

\subsection{Fuzzy inference system}
To achieve an interpretable fuzzy rule-based classifier (FRBC), we set the maximum number of rules at $15$ and limit the number of antecedents to $3$. However, the number of rules is further reduced based on a quality metric. We also define $3$ linguistic labels for the fuzzy partitions—low, medium, and high—which are intuitive and straightforward to interpret.

These partitions are derived empirically from the available data. The "low" partition has its maximum membership value from the minimum data value up to the 0.2 quantile, decreasing to the 0.5 quantile. The "medium" partition starts at the 0.2 quantile, peaks at the 0.5 quantile, and decreases until the 0.8 quantile. Similarly, the "high" partition spans the 0.5 to 1 quantiles, peaking at the 0.8 quantiles. When using interval type-2 fuzzy sets, the lower membership function is capped at a maximum degree of $0.8$.
For the inference process, we adopt the scheme implemented in the Ex-Fuzzy library \cite{fumanalex2024}. The degree of matching for each rule is quantified using its dominance score, $ds_r$, and its firing strength. The firing strength is computed as the product of the truth degrees of all rule antecedents. The dominance score is defined as \cite{kiani2022temporal, andreu2021explainable}:
\begin{itemize}
	\item The fuzzy support of the rule:
	\begin{equation} \label{eq:support}
		s_r = \frac{\sum_{\mathbf{x} \in 				Cons_r} w_r (\mathbf{x})}{|\mathbf{x} \in 				Cons_r|}.
	\end{equation}
	\item The fuzzy confidence of the rule:
	\begin{equation}
		c_r = \frac{\sum_{\mathbf{x} \in Cons_r} w_r (\mathbf{x})}{{\sum_{\mathbf{x} \in Cons_r }\sum_{r'=1}^{R}{w_{r'}(\mathbf{x})}}}.
	\end{equation}
	
\end{itemize}   

\noindent Where $w_r(\mathbf{x})$ is the firing strength of the rule $r$ for the sample $\mathbf{x}$. Also, $\mathbf{x} \in Cons_r$ is the set of observations whose ground-truth class corresponds to the rule class consequent. $R$ is the set of all rules. 

Finally, we compute the association degree, $as_r(x)$, using $w_r(\mathbf{x})$ and $ds_r$:
\begin{equation}
	as_r(x) = w_r(\mathbf{x}) * ds_r.
\end{equation}

Each sample is classified according to the consequent class of the rule with the highest association degree for that sample.

For our experimentation, we used the genetic algorithm to optimize the antecedents and consequencess of each rule using the previously computed partitions. The metric to optimize is the Matthew correlation coefficient (MCC):
    \begin{equation} \label{eq:MCC}
	MCC = \frac{(TP \times TN) - (FP \times FN)}{\sqrt{(TP + FP)(TP + FN)(TN + FP)(TN + FN)}}
\end{equation}

where TP is true positive, TN means true negative, FP is false positive, and FN is false negative.

We use a cross-validation method to empirically estimate non-conformal scores \cite
{vovk2015cross}. This consists of dividing the data into K-folds and then training a model using K-1 folds and the remaining fold for calibration. We aggregate the non-conformity scores across all folds, which is more data efficient as it uses each point for training and calibration.

\subsection{Adapting the fitting function to conformal learning}
Since we are conformalizing the predictor, we can also adapt the fitness function to optimize of the sets of classes predicted. Given the non-coformal score $s_i=1-\hat{f}(X_i)_Y$ (or its interval-valued equivalent), both Eq. (\ref{eq:conformal_pred}) and Eq. (\ref{eq:conformal_pred_iv}) already make sure that we are obtaining the smallest prediction set on average \cite{balasubramanian2014conformal}. However, we can change the fitness function so that the classifier is aware of this and we can exploit it further.

Usually, we use the MCC function as the only fitness function, which is good for singleton predictions. However, we can incorporate an additional term that warrantees that the rules in the FRBC have the best non-conformal scores as well for the non-true classes:

\begin{equation} \label{eq:penalizing_fitting}
    L_2 = \sum_r^R \sum_{i=1}^{N} t_{i,r},
\end{equation}

where $Y'_i$ is the incorrect classes of a sample $i$, and:

\begin{equation}
    t_{i,r} = \sum_{y \in Y'_i} r(X_i)_y.
\end{equation}

\noindent Since $r(X_i)$ is the same for all classes, we can see this equation, and thus, Eq. (\ref{eq:penalizing_fitting}), as penalizing the support of the rule. Following the example in Figure \ref{fig:rulewiseconformal}, this equation would penalize R1 and R2, which are the rules that are firing most and causing more confusion with other rules.

\section{Experimentation} \label{sec:exp}

The accuracy results obtained for each classifier is detailed in Table \ref{tab:exp_results}. Here, we show the average of the results for five different classifiers trained.

There are different metrics to study: 

\begin{itemize}
    \item The size of the predictions at the desired level of significance. Conformal prediction aims to produce prediction sets that satisfy a pre-specified coverage guarantee with a plausibility leve of $1 - \alpha$ (e.g., 95\%). The size of the prediction set determines whether the method achieves its coverage guarantees. If the prediction set is too small, it might not cover the true label with the desired frequency, leading to invalid predictions. Although larger prediction sets might ensure that the true label is included more often, they could be overly inclusive, making the prediction less informative or useful.
    \item The number of predictions that are not empty and which significance level should we put to have at least one prediction for each sample in the dataset.
\end{itemize}

\begin{figure}[ht]
    \centering
    \includegraphics[width=\linewidth]{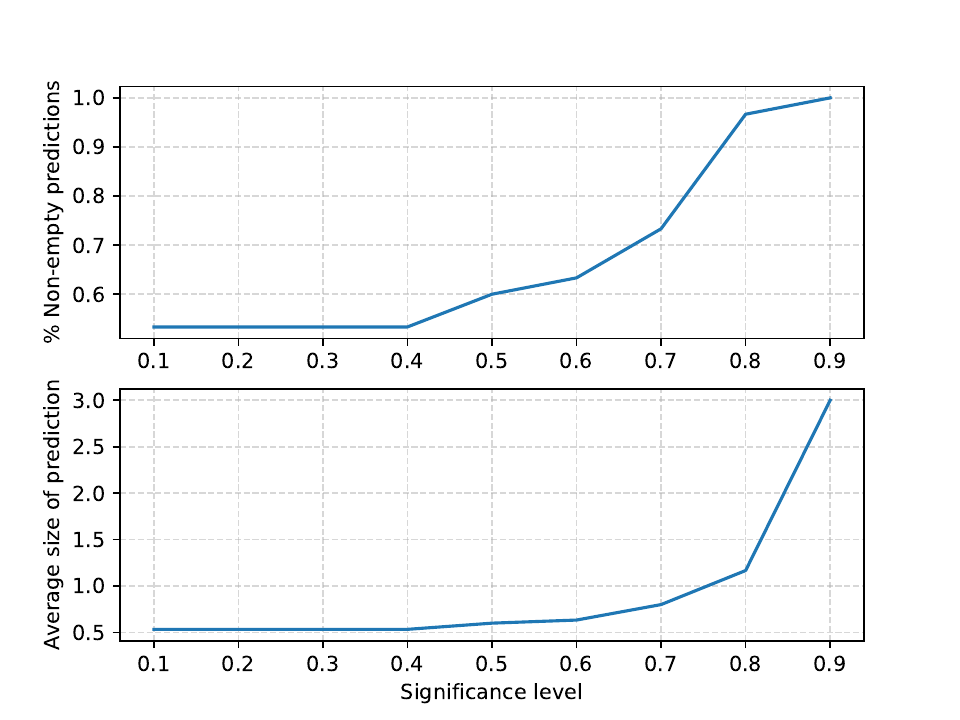}
    \caption{Evolution of prediction size and percentage of valid predictions using IV-T2 fuzzy sets in the Iris dataset.}
    \label{fig:sample_iris}
\end{figure}

Figure \ref{fig:sample_iris} shows an example of the evolution of the prediction size and the proportion of non-empty predictions according to the significance level for T1 and IV-T2 datasets, where we can see that using a big significance level can very easily result in having trivial results, and thus, the importance on correctly setting it beforehand.

\begin{figure}[ht]
    \centering
    \includegraphics[width=\linewidth]{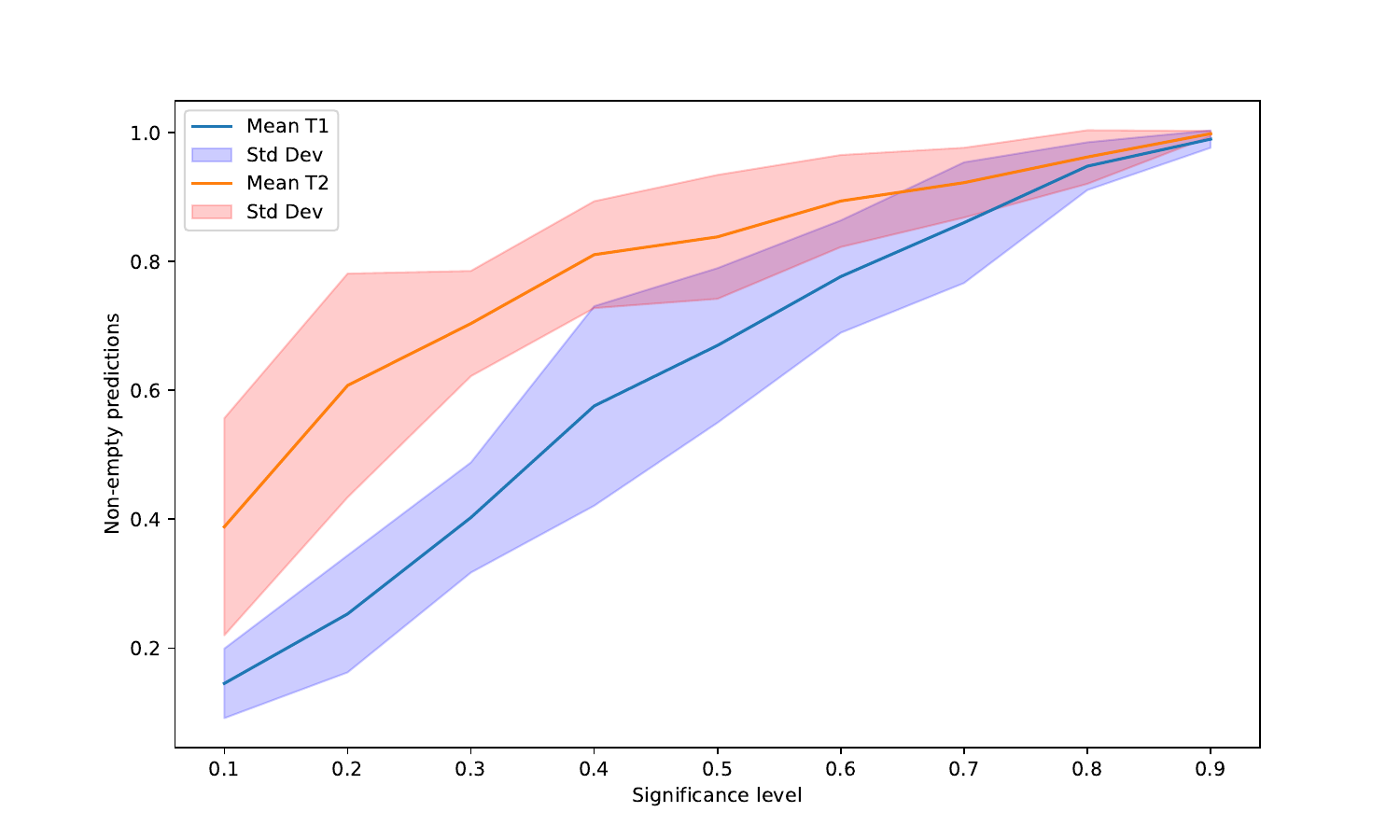} \\
    \includegraphics[width=\linewidth]{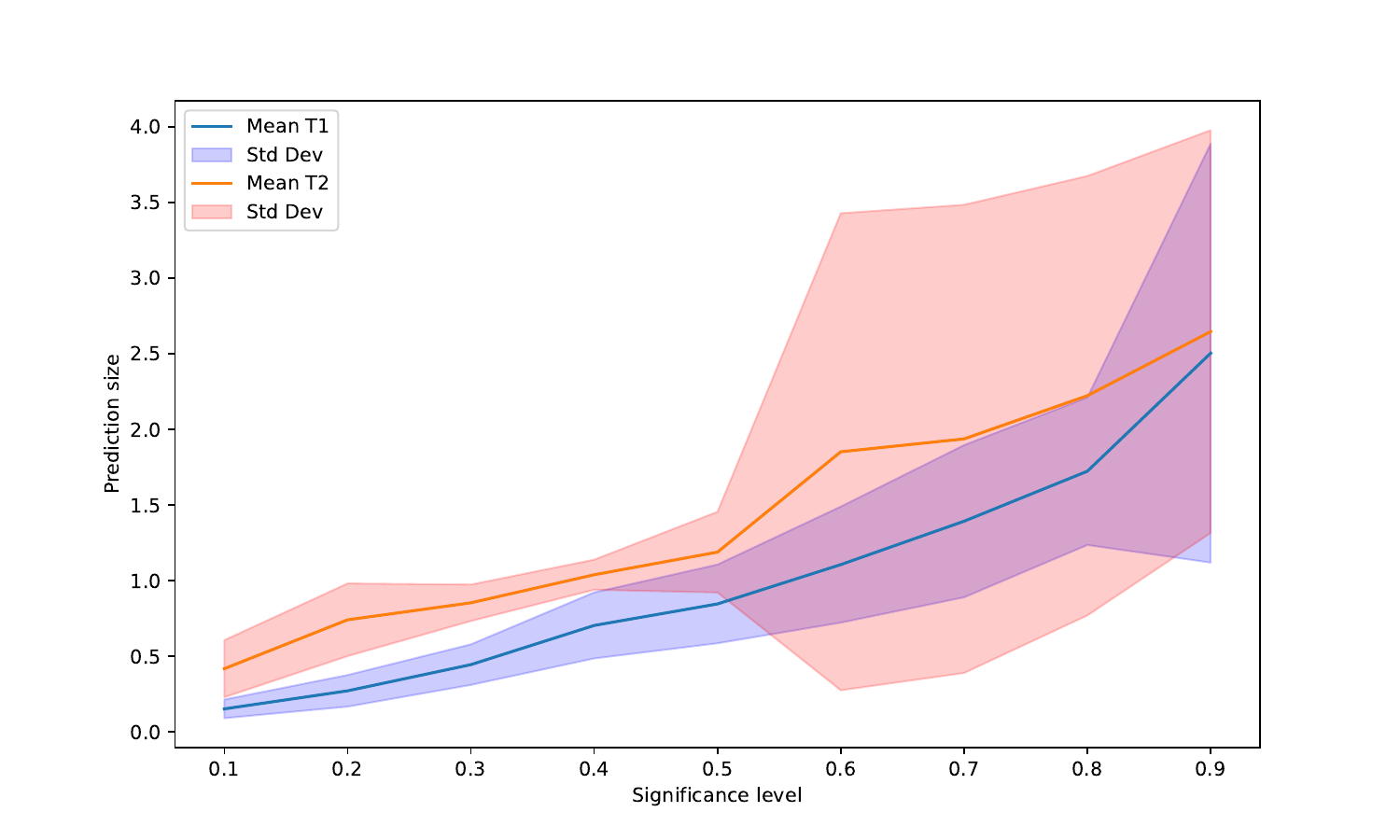}
    \caption{Prediction size and non-empty predictions depending on the significance level desired.}
    \label{fig:classwise-res}
\end{figure}

In Figure \ref{fig:classwise-res} we show the results for the different levels of alpha of both fuzzy sets across different levels of significance. We show the lowest level of significance in which we find the classifier with non-empty predictions.

\begin{figure}[ht]
    \centering
     \includegraphics[width=\linewidth]{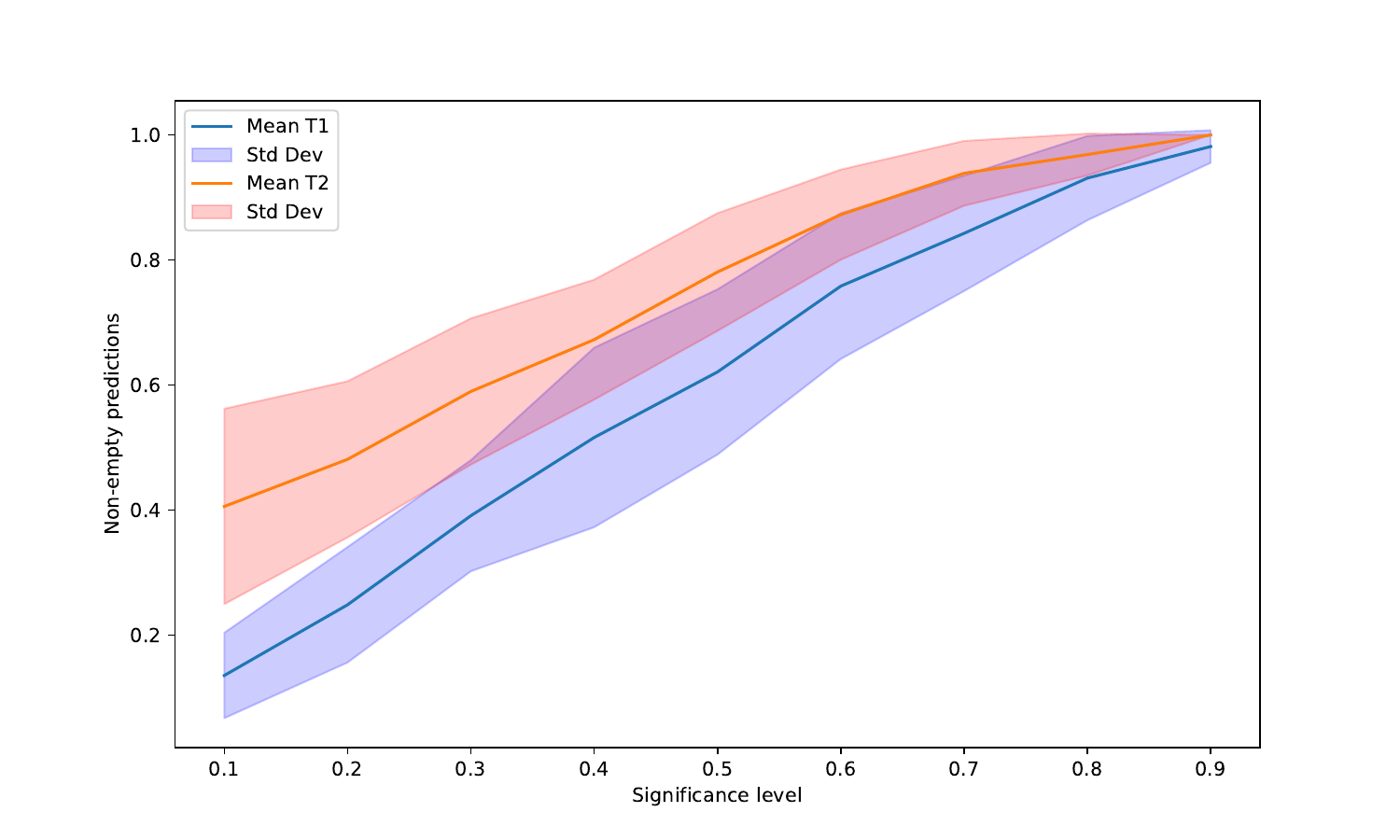} \\
    \includegraphics[width=\linewidth]{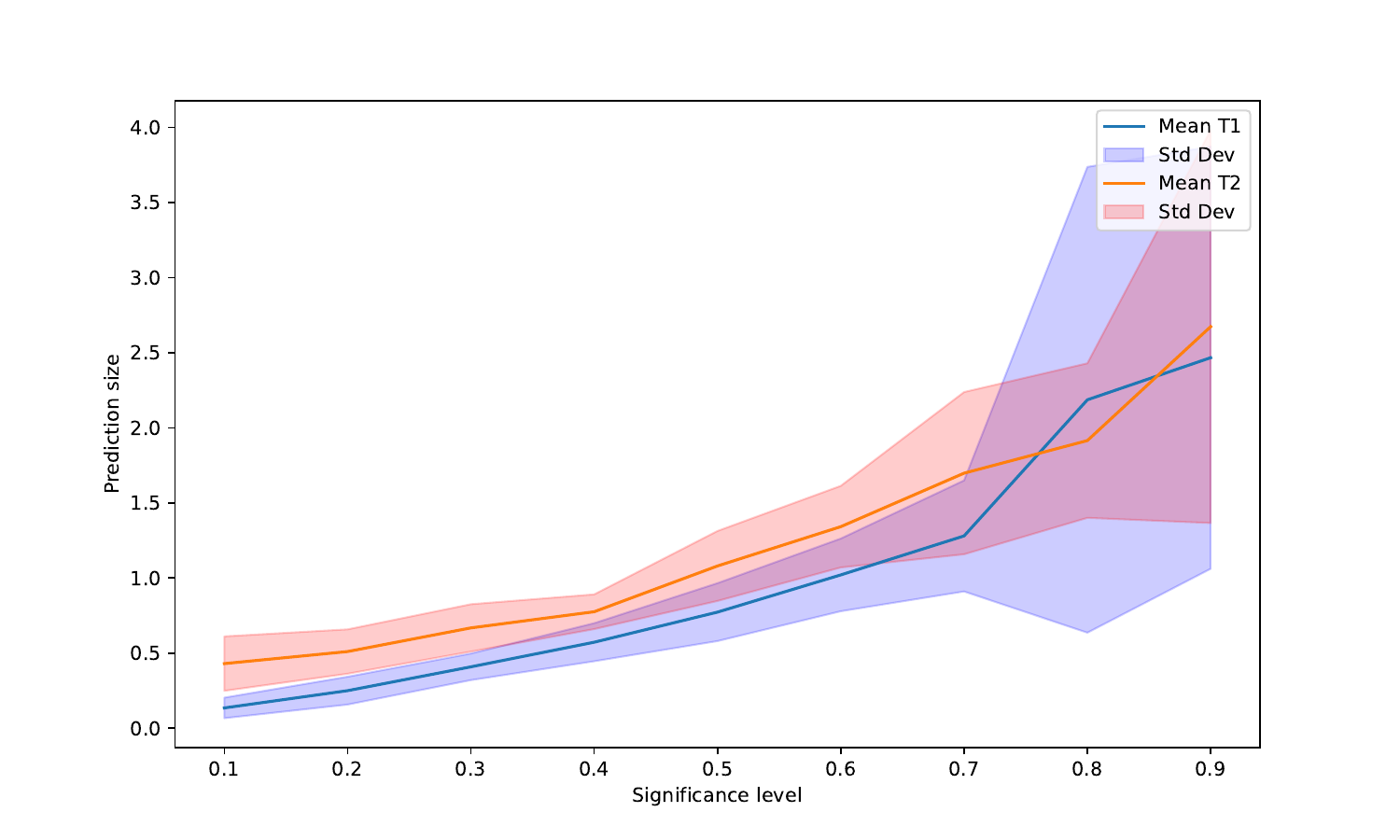}
    \caption{Prediction size and non-empty predictions depending on the significance level desired using a custom fitness function to reduce the average prediction size.}
    \label{fig:classwise-resFIT}
\end{figure}

\begin{figure}[ht]
    \centering
    \includegraphics[width=\linewidth]{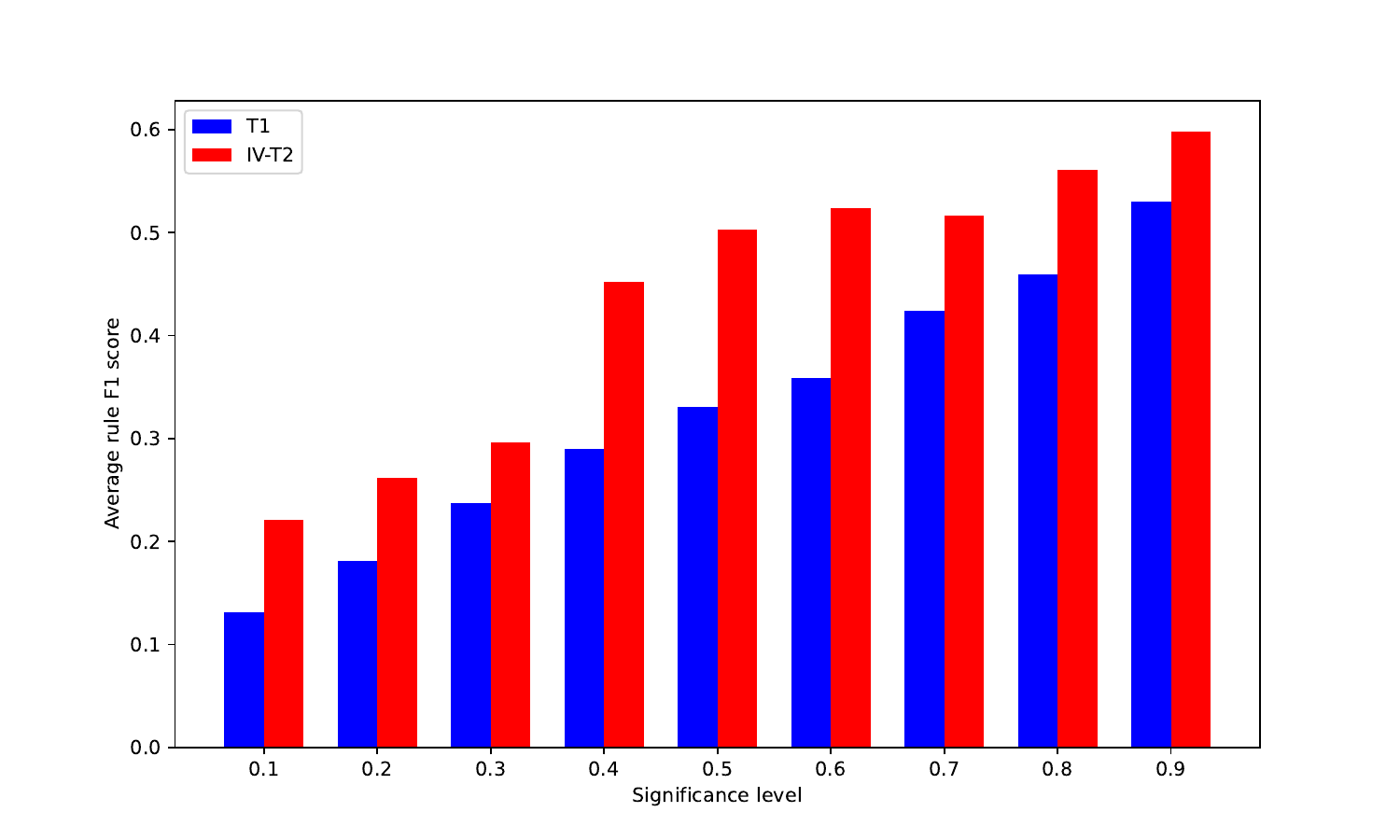}
    \caption{Average F1 score of all the rules for T1 and IV-T2 fuzzy sets.}
    \label{fig:rulewise-res}
\end{figure}

Then, in Figure \ref{fig:classwise-resFIT} and Fig. \ref{fig:rulewise-res}, we show the same results, classwise and rulewise, when using a Laplacian term in the fitness function that incorporates Eq.(\ref{eq:penalizing_fitting}) with a $0.01$ multiplier.

For rule-wise learning, we show in Figure \ref{fig:rulewise-res} the evolution in the average F1 score for each rule according to the significance level desired. 

\begin{table}[ht]
    \centering
    \caption{Accuracy results for 5 fold validation for each dataset}
    \begin{tabular}{c|cc}
    \toprule
         Dataset & T1 & IV-T2 \\
         \midrule
         iris & $96.66\pm0.03$ & $92.22\pm1.01$\\
         pima & $72.56\pm0.38$ & $64.27\pm4.76$\\
         glass &$44.39\pm0.30$& $45.79\pm2.75$\\
         haberman &$76.68\pm 1.10$& $74.94\pm1.34$\\
         ionoshpere &$81.19\pm0.71$& $83.00\pm2.08$\\
         wine &$71.91\pm10.84$&$71.53\pm2.71$\\
         balance &$54.34\pm1.43$& $52.58\pm2.41$\\
         heart &$71.35\pm1.06$ & $71.97\pm1.27$\\
         \bottomrule
    \end{tabular}
    \label{tab:exp_results}
\end{table}

\subsection{Discussion}
In the experimentation section we have tested different configurations of FRBC classifiers, T1 and T2 fuzzy sets, standard and custom fitness function; looking for a good trade-off between performance and prediction size using conformal learning. 

We found that IV-T2 lacked performance compared to equivalent T1 classifiers in some cases. This could be due to the fact that intervals require more parameters to fine tune, and thus training for them could benefit from larger training times. However, for the sake of fairness, we have shown the results for both fuzzy sets using the same number of generations for genetic fine-tuning.

Regarding the prediction size, which would ideally be as low as possible, we found again that IV-T2 offered marginally worse results. However, the percentage of non-empty predictions was better for IV-T2 fuzzy sets, which might make them more effective in many real-world applications where the classifier should at least predict one class. The standard deviation for prediction size is also considerable for IV-T2 systems, which reinforces the idea that the genetic fine tuning requires more computation than T1 systems to achieve stable results.

When using a custom fitness function to reduce the size of the prediction, we found that IV-T2 really benefited from it. It reduced the prediction size and standard deviation for large significance values without negatively affecting the number of non-empty predictions. However, T1 FRBCs showed no significant improvement in this regard.

Rulewise results, we found that for both T1 and IV-T2 fuzzy sets the F1 score increases alongside the the significance level required. Further exploration of this result showed that this happened because most rules received an increase in recall without hurting precision at the higher levels of significance. We also found that the rules in the IV-T2 classifiers consistently achieved better F1 score values than those using T1 fuzzy sets.

\section{Conclusions} \label{sec:conclusions}
In this paper, we show how conformal learning can be integrated with fuzzy rule-based systems to enhance uncertainty quantification in machine learning classification tasks. Our approach also incorporates interval-valued Type 2 fuzzy sets, which offer an additional layer to handle epistemic uncertainty.

Our main results show:
\begin{itemize}
\item While IV-T2 fuzzy systems provide better uncertainty quantification and non-empty prediction coverage, they often require more computational resources and fine-tuning compared to Type 1 (T1) fuzzy systems. However, they also generate more reliable rules.

\item Adapting fitness functions for conformal prediction yielded significant improvements in prediction set size and standard deviation, particularly for IV-T2 fuzzy systems.
\end{itemize}

Our future work could explore the use of General Type 2 and additional uncertainty quantification metrics. Additionally, we want to introduce fuzzy classifiers into computer vision frameworks, which offer larger and more diverse datasets and where conformal learning has also proven to be very beneficial.

\section{Acknowledgements}
Javier Fumanal-Idocin research has been supported by the European Union and the University of Essex under a Marie Sklodowska-Curie YUFE4postdoc action.

\bibliographystyle{IEEEtran}
\bibliography{conference_101719}

\end{document}